\documentclass[conference]{IEEEtran}
\IEEEoverridecommandlockouts
\usepackage{cite}
\usepackage{amsmath,amssymb,amsfonts}
\usepackage{algorithmic}
\usepackage{graphicx}
\usepackage{textcomp}
\usepackage{booktabs}
\usepackage{multirow}
\usepackage{xcolor}
\def\BibTeX{{\rm B\kern-.05em{\sc i\kern-.025em b}\kern-.08em
    T\kern-.1667em\lower.7ex\hbox{E}\kern-.125emX}}

\usepackage{listings}
\usepackage{xcolor}
\lstdefinestyle{pyStyle}{
  language=Python,
  basicstyle=\ttfamily\small,
  numbers=left,
  numberstyle=\tiny,
  frame=lines,
  numbersep=4pt,
  breaklines=true,
  showstringspaces=false,  
  columns=flexible,        
  keepspaces=true,         
  keywordstyle=\color{blue},
  commentstyle=\color{gray!70!black},
  stringstyle=\color{red!60!black},
  tabsize=2,
}
\usepackage[hidelinks]{hyperref}
\begin{document}


\title{Chain-of-Thought Reasoning with Large Language Models for Clinical Alzheimer’s Disease Assessment and Diagnosis}

\author{%
\centering
\begin{tabular}{ccc}
\begin{minipage}[t]{0.30\textwidth}\centering\vspace{0pt}
Tongze Zhang ~\href{https://orcid.org/0000-0002-3375-7136}{\includegraphics[scale=0.06]{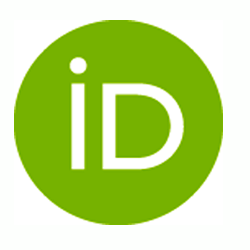}}\\
\textit{Stevens Institute of Technology}\\
Hoboken, New Jersey
\end{minipage}
&
\begin{minipage}[t]{0.30\textwidth}\centering\vspace{0pt}
Jun-En Ding ~\href{https://orcid.org/0000-0002-1233-138X}{\includegraphics[scale=0.06]{graph/orcid.png}}\\
\textit{Stevens Institute of Technology}\\
Hoboken, New Jersey
\end{minipage}
&
\begin{minipage}[t]{0.30\textwidth}\centering\vspace{0pt}
Melik Ozolcer ~\href{https://orcid.org/0000-0003-4251-0204}{\includegraphics[scale=0.06]{graph/orcid.png}}\\
\textit{Stevens Institute of Technology}\\
Hoboken, New Jersey
\end{minipage}
\\[0.9ex]

\begin{minipage}[t]{0.30\textwidth}\centering\vspace{0pt}
Fang-Ming Hung ~\href{https://orcid.org/0000-0003-3501-5459}{\includegraphics[scale=0.06]{graph/orcid.png}}\\
\textit{Surgical Trauma Intensive Care Unit}\\
Far Eastern Memorial Hospital
\end{minipage}
&
\begin{minipage}[t]{0.30\textwidth}\centering\vspace{0pt}
Albert Chih-Chieh Yang ~\href{https://orcid.org/0000-0003-2794-9649}{\includegraphics[scale=0.06]{graph/orcid.png}}\\
\textit{Institute of Brain Science}\\
National Yang Ming Chiao Tung University
\end{minipage}
&
\begin{minipage}[t]{0.30\textwidth}\centering\vspace{0pt}
Feng Liu ~\href{https://orcid.org/0000-0002-5225-8199}{\includegraphics[scale=0.06]{graph/orcid.png}}\\
\textit{Stevens Institute of Technology}\\
Hoboken, New Jersey
\end{minipage}
\\[1.0ex]

\multicolumn{3}{c}{%
\begin{minipage}[t]{0.34\textwidth}\centering\vspace{0pt}
Yi-Rou Ji ~{\includegraphics[scale=0.06]{graph/orcid.png}}\\
\textit{Surgical Trauma Intensive Care Unit}\\
National Yang Ming Chiao Tung University
\end{minipage}
\hspace{2.2em}
\begin{minipage}[t]{0.34\textwidth}\centering\vspace{0pt}
Sang Won Bae*~\href{https://orcid.org/0000-0002-2047-1358}{\includegraphics[scale=0.06]{graph/orcid.png}}\\
\textit{Stevens Institute of Technology}\\
Hoboken, New Jersey
\end{minipage}
}
\end{tabular}
}

\maketitle

\begin{abstract}
Alzheimer's disease (AD) has become a prevalent neurodegenerative disease worldwide. Traditional diagnosis still relies heavily on medical imaging and clinical assessment by physicians, which is often time-consuming and resource-intensive in terms of both human expertise and healthcare resources. In recent years, large language models (LLMs) have been increasingly applied to the medical field using electronic health records (EHRs), yet their application in Alzheimer's disease assessment remains limited, particularly given that AD involves complex multifactorial etiologies that are difficult to observe directly through imaging modalities. In this work, we propose leveraging LLMs to perform Chain-of-Thought (CoT) reasoning on patients' clinical EHRs. Unlike direct fine-tuning of LLMs on EHR data for AD classification, our approach utilizes LLM-generated CoT reasoning paths to provide the model with explicit diagnostic rationale for AD assessment, followed by structured CoT-based predictions. This pipeline not only enhances the model's ability to diagnose intrinsically complex factors but also improves the interpretability of the prediction process across different stages of AD progression. Experimental results demonstrate that the proposed CoT-based diagnostic framework significantly enhances stability and diagnostic performance across multiple CDR grading tasks, achieving up to a 15\% improvement in F1 score compared to the zero-shot baseline method.
\end{abstract}

\begin{IEEEkeywords}
Alzheimer’s Disease, Large Language Models, Chain-of-Thought Reasoning, Clinical Decision Support, Electronic Health Records, Neurodegenerative Disorders
\end{IEEEkeywords}

\section{INTRODUCTION}
Alzheimer’s disease (AD) is the most prevalent neurodegenerative disorder worldwide, posing a growing challenge to healthcare systems and patients’ quality of life. The rising global prevalence of AD creates an urgent need for accurate, scalable, and resource-efficient diagnostic tools. Traditionally, standard diagnostic protocols remain heavily reliant on a combination of time-consuming and costly methods, including advanced medical imaging modalities such as PET and structural MRI, alongside clinical assessments by specialized physicians \cite{ding2025variational}\cite{alia2024daily}. This high dependence on human expertise and expensive infrastructure limits accessibility.

\textcolor{black}{Alzheimer's disease is characterized by progressive impairment of cognitive and functional behavior, encompassing memory, orientation, judgement, and the ability to perform daily activities. Such behavioral changes are typically documented within the clinical narratives of electronic health records (EHRs), where clinicians describe patients' functional status, behavioral symptoms, and activities of daily living using natural language. The Clinical Dementia Rating (CDR) serves as a standardized clinical assessment tool \cite{li2025care}, synthesizing multidimensional cognitive and functional behavioral performance.}

In recent years, the integration of large language models (LLMs) with EHR has transformed predictive analytics in medicine. LLMs are increasingly applied to diverse tasks ranging from clinical document summarization to patient outcome prediction. However, their direct application to inherently multifactorial complex diseases like AD remains constrained. Despite advances in LLMs for understanding medical texts, their application in diagnostic decision-making remains somewhat limited. Traditional LLM fine-tuning often produces black-box classifiers with limited interpretability and traceability, constraining their clinical adoption \cite{du2025testing}. 

To bridge this critical gap in explainability and complex reasoning, we propose a novel diagnostic workflow leveraging Chain-of-Thought (CoT) reasoning within LLMs for AD assessment based on comprehensive clinical EHR data \cite{lucas2024reasoning}. Our approach achieves this by explicitly generating intermediate diagnostic reasoning. These LLM-generated CoT paths are designed to mimic the explicit, stepwise logical reasoning process employed by clinical experts, transforming heterogeneous EHR features into structured, verifiable explanations before arriving at a final prediction. By integrating this explicit reasoning layer, the model demonstrates enhanced capability in handling the intrinsically complex factors defining AD pathogenesis and progression. The results demonstrate that incorporating structured reasoning significantly enhances diagnostic consistency and interpretability while maintaining performance. This research highlights the potential of CoT-enhanced large language models to bridge automated prediction with interpretable clinical reasoning, laying the foundation for trustworthy AI-assisted diagnostic systems in neurodegenerative disease research. Existing research primarily relies on longitudinal clinical records to predict whether or when Alzheimer’s disease will occur. In contrast, this study focuses on clinical staging through the grading of CDR within current electronic health record texts, emphasizing interpretable, real-time assessments to support clinical decision-making. \textcolor{black}{This work presents a structured, multi-stage Chain-of-Thought reasoning framework for interpretable and stable behavioral assessment from clinical narratives using large language models.}

To address the current lack of interpretability and reliability in language models for Alzheimer's disease (AD) diagnosis, this study primarily explores the following three key questions:




1. Will LLM reliably identify subtle CDR grading differences from unstructured EHRs?

2. Can incorporating CoT reasoning enhance interpretability and credibility?

3. Can multi-stage reasoning structures reduce performance fluctuations across different CDR grading tasks, thereby enhancing prediction consistency?

\section{Related Work}
\subsection{Traditional Approaches to Alzheimer’s Disease Diagnosis}
The diagnosis of Alzheimer's disease (AD) has traditionally relied upon neuropsychological assessments, imaging examinations (such as MRI and PET), and cerebrospinal fluid biomarkers \cite{ding2025variational}. Whilst these methods possess clinical value, they are costly, invasive, and difficult to deploy at scale for screening purposes \cite{fikry2022modelling}. In recent years, machine learning and deep learning approaches have been applied to neuroimaging data for automated AD detection and disease progression prediction. For instance, convolutional neural networks (CNNs) applied to MRI or PET scans have achieved high accuracy in distinguishing mild cognitive impairment (MCI) from AD \cite{wang2025flexible}\cite{dao2025curenet}. However, these models often exhibit “black box” behaviour, lacking transparent reasoning pathways. Complementing imaging research, natural language processing (NLP) methods have also been applied to EHR texts: analysing narrative clinical notes, cognitive test summaries, and EHR time-series data can detect cognitive decline and predict conversion to AD \cite{amini2024prediction}. However, many NLP-based systems rely on bag-of-words models, sequence classification, or fine-tuned Transformer models. While these approaches capture surface-level patterns, they fail to explicitly model clinical reasoning processes or achieve interpretability. Basic clinical staging and biomarker frameworks further underscore the importance of mild cognitive impairment as the biological precursor stage of Alzheimer's disease. These findings have guided the design of recent machine learning models \cite{petersen2011mild}\cite{jack2018nia}. Systematic reviews and meta-analyses of MRI/PET-based machine learning systems have reconfirmed performance improvements while simultaneously demonstrating comparability and data leakage risks within convolutional neural network pipelines \cite{wen2020convolutional}\cite{battineni2024machine}.

\subsection{Large Language Models in Medical Text Understanding}

The advent of LLMs, such as ClinicalBERT, LLaMA, and domain-adapted models, has significantly advanced medical natural language processing. Recent reviews on LLMs in healthcare highlight their broad applicability across tasks including entity extraction, report summarisation, and outcome prediction \cite{nazi2024large}. Research indicates that when applied to clinical texts, LLMs surpass earlier smaller models, demonstrating potential for diagnostic reasoning \cite{lievin2024can}. Despite these advances, however, many LLM-based clinical systems suffer from opacity issues: they provide only classification results or recommendations without displaying intermediate reasoning steps, thereby limiting clinician trust and interpretability.

Recent clinical text studies on LLMs have primarily focused on information extraction and classification tasks, typically providing only classification results or evidence snippets without systematic generative clinical reasoning chains. Simultaneously, many validation studies suffer from small data scales (e.g., evaluating a limited number of manually reviewed cases) or are constrained by single-center, controlled settings, making it difficult to cover complex real-world clinical contexts \cite{zhang2024evaluating}. Beyond early domain models such as ClinicalBERT \cite{huang2019clinicalbert}, recent studies indicate that LLMs with safety alignment mechanisms and instruction-based fine-tuning—such as the Med-PaLM series—can encode extensive clinical knowledge yet still underperform compared to clinicians in fine-grained tasks \cite{singhal2023large}\cite{singhal2025toward}. Randomized trials also indicate that providing large language model assistance does not uniformly enhance physicians' diagnostic reasoning capabilities, underscoring the necessity of establishing transparent, auditable workflows \cite{goh2024large}. Our approach addresses this by designing an auditable multi-stage reasoning template that generates clinically readable intermediate conclusions and final consolidated opinions for each record. This aims to enhance interpretability and clinical utility while maintaining performance.

\subsection{Chain-of-Thought (CoT) Reasoning and Explainable AI in Medicine}

CoT prompts represent a recently developed methodology within large language model research, requiring models to articulate intermediate reasoning steps rather than directly outputting answers \cite{wei2022chain}. This stepwise reasoning approach has demonstrated improved performance on arithmetic and common-sense benchmarks. Within the medical domain, CoT has been explored to enhance the interpretability, auditability, and coordination with clinicians of large language model decision-making \cite{miao2024chain}. For instance, a renal disease diagnosis study demonstrated that CoT prompts enable LLMs to expose decision pathways and facilitate error tracing \cite{miao2024chain}. Recent structured clinical reasoning prompts—incorporating differential reasoning, analytical reasoning, and Bayesian inference frameworks—further expand LLM potential in medical tasks \cite{sonoda2025structured}. Despite these advances, CoT reasoning applications in neurodegenerative diseases like Alzheimer's remain in their exploratory infancy. Recent work proposes a CoT-based Alzheimer's disease classification method, achieving approximately 16.7\% performance improvement through supervised fine-tuning with reasoning cues \cite{park2025reasoning}. Moreover, emerging research employing multi-agent large language model frameworks for early Alzheimer's detection from longitudinal clinical records highlights a trend towards simulating expert consultation workflows \cite{li2025care}. These research gaps motivated our present work: we propose a multi-stage CoT diagnostic system specifically designed for Alzheimer's assessment. By integrating structured reasoning with consensus-building mechanisms, it significantly enhances interpretability, consistency, and clinical relevance.

Strengthening CoT by sampling diverse reasoning paths and aggregating consensus provides a principled approach to reducing variability in clinical reasoning chains \cite{wang2022self}. CoT-based diagnostic reasoning has also demonstrated improved interpretability in controlled clinical benchmarks \cite{savage2024diagnostic}, while healthcare multi-agent frameworks based on longitudinal medical records have proven the feasibility of simulating collaborative team-based diagnosis and treatment processes \cite{li2025care}. Research on CoT and structured diagnostic prompts in medical settings has been explored, but most work remains focused on promoting accurate classification rather than generating comprehensive, verifiable explanatory texts based on clinical semantics \cite{savage2024diagnostic}. Additionally, existing studies often evaluate performance on limited-scale tasks or benchmarks. Our design addresses the challenges of small-sample and out-of-domain generalization while ensuring interpretability.

\section{Method}

This study aims to construct an interpretable automated CDR grading system based on a two-stage experimental framework. The first stage focuses on establishing multiple high-performance classification baselines to provide impartial performance benchmarks. The second stage introduces the core innovation that a CoT Alzheimer’s Disease Diagnostic system to address the transparency issues of traditional black-box models in complex clinical reasoning.

\subsection{Data Preprocessing}

\begin{table*}[htbp]
\centering
\caption{Representative real-world CDR-labeled clinical records showing the complexity of Subjective (S) and Assessment (A) texts.}
\begin{tabular}{p{0.8cm}p{7.2cm}p{7.2cm}}
\hline
\textbf{CDR} & \textbf{Subjective Note (S)} & \textbf{Assessment (A)} \\
\hline
0.5 & “insidious onset with progressive poor memory forgets conversation details but able to manage home affairs; occasional confusion reported by spouse." & “Autistic thinking (+) vague responses; mild impairment in orientation and memory domains; functional independence maintained." \\
\hline
1.0 & “WITH daughter deterioration multiple complaints of forgetfulness, misplacing items, and poor concentration; sometimes fails to find way home." & “Suspect depression (treated at Psyche) reports, mild to moderate decline, impaired attention span, partial insight preserved." \\
\hline
2.0 & “progressive for years with poor memory Forgetfulness noted by family members and reduced self-care; occasional urinary incontinence reported." & “(favor) in progression with incontinence? cognitive decline with temporal disorientation, impaired judgment, and dependency for daily activities." \\
\hline
3.0 & “request application for disability certificate due to long-term confusion, unable to recognize relatives; total dependence for self-care." & “Right PCA territory infarct (onset: )(TOAST type: ) consistent with severe dementia picture; bed bound, nonverbal, requires full assistance." \\
\hline
\end{tabular}
\label{tab:cdr_real_samples}
\end{table*}

The raw data used in this study originated from a five year clinical dataset containing patients' longitudinal EHR, primarily comprising the patient's Subject (S) and clinical diagnostic Assessment (A) fields. To ensure data quality and analytical validity, we first executed a rigorous data cleaning process. Since multiple records might exist for the same patient across different time points, we utilized unique medical record identifiers and performed deduplication based on the longest text length principle. The initial dataset contained 745 raw patient records. Following preprocessing, all samples with empty assessment (A) fields were excluded, yielding 698 distinct and complete patient records suitable for subsequent modeling. The CDR labels in our dataset cover four clinically recognized Alzheimer’s disease severity levels: 0.5, 1.0, 2.0, and 3.0, corresponding to very mild, mild, moderate, and severe dementia, respectively. Each stage exhibits distinct functional characteristics: CDR 0.5 patients typically show mild memory deficits while maintaining daily independence; CDR 1.0 patients exhibit more pronounced cognitive and functional impairments; those with 2.0 require assistance for daily tasks; and those with 3.0 exhibit severe disorientation and complete dependence on caregivers.



 This experimental design employs a systematic one-versus-one binary classification strategy, decomposing the complex multi-class CDR problem into a series of more specific and controllable diagnostic subtasks. The CDR scale is a widely recognized clinical standard for quantifying the degree of cognitive decline across multifunctional domains such as memory, orientation, judgment, and self-care. 
 
To ensure consistency in assessment across different levels of disease severity, we constructed four binary classification subsets from the final dataset: 0.5 vs 1.0, 0.5 vs 2.0, 0.5 vs 3.0, and 1.0 vs 3.0. The 0.5 vs 1.0 subset contained 429 patient records, the 0.5 vs 2.0 subset contained 340 patient records, the 0.5 vs 3.0 subset contained 263 patient records, and the 1.0 vs 3.0 subset contained 358 patient records. Each subset was independently processed through the proposed CoT inference framework to evaluate diagnostic performance across varying degrees of cognitive impairment. Representative samples of these records are presented in Table \ref{tab:cdr_real_samples}. Each binary experiment focused on adjacent or clinically significant stage differences, enabling the system to precisely identify diagnostic boundaries. The Clinical Dementia Rating (CDR) itself reflects the progressive deterioration of multidimensional functions in Alzheimer's disease patients, including memory, orientation, judgment, and self-care abilities. Therefore, the 0.5 vs 1.0 and 0.5 vs 2.0 comparisons assess whether the model can detect subtle early-stage differences in disease progression, while the 0.5 vs 3.0 and 1.0 vs 3.0 comparisons validate the model's robustness across scenarios with significant functional gaps. Compared to directly constructing a multi-class classification model, decomposing the task into binary sub-tasks corresponding to clinically meaningful decision boundaries reduces label ambiguity and enhances interpretability. This grouping approach thus possesses clear medical significance while providing the model with well-defined, verifiable classification objectives. This enables systematic evaluation of reasoning stability and discriminative capability across different levels of cognitive decline.

 
For each paired task, the system dynamically extracts patient records corresponding to the two target CDR grades from the preprocessed dataset. These subsets are then divided into training (80\%) and test (20\%) sets, ensuring balanced representation across severity levels. This design not only simplifies the model's decision space but also enables targeted evaluation of its reasoning stability and generalization capabilities under diverse clinical contrast conditions. Through this approach, the study systematically explores how the proposed reasoning framework adapts to varying degrees of cognitive impairment, revealing its diagnostic interpretability and sensitivity to disease progression.



\subsection{CoT Diagnostic System: Independent CoT Generation and Reasoning Validation}

\begin{figure*}
    \centering
    \includegraphics[width=1\linewidth]{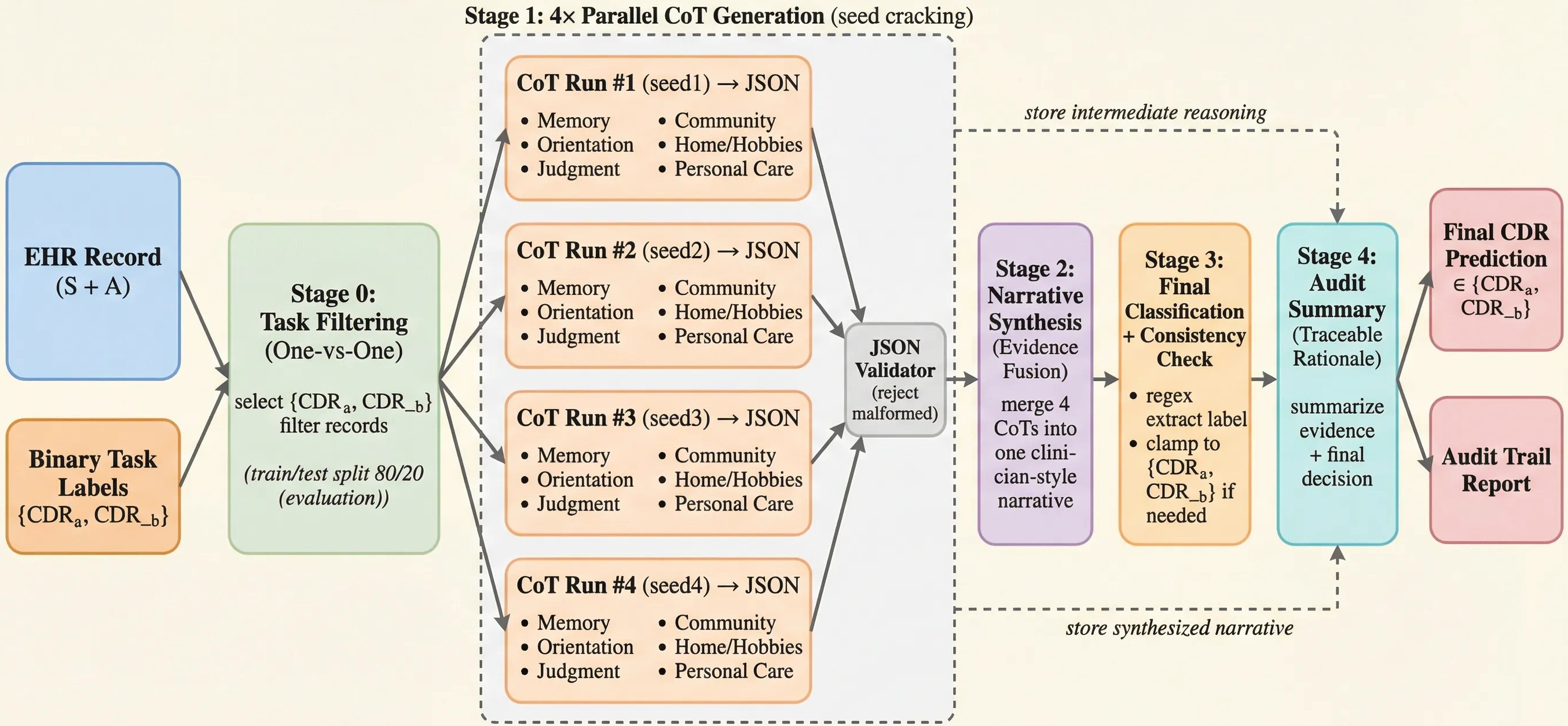}
    \caption{Four-Step Pipeline for Binary CDR Classification}
    \label{fig:placeholder}
\end{figure*}

\begin{lstlisting}[style=pyStyle, caption={Chain-of-Thought Prompting Structure for CDR Reasoning}, label={lst:prompt_structure}, breaklines=true, breakatwhitespace=true, basicstyle=\ttfamily\small]
System prompt:
"You are an experienced neurologist specializing in Alzheimer's disease.
Respond professionally in English and output valid JSON."

User prompt:
# Task: Generate a full clinical analysis from a subjective note
Input: "Patient reports occasional disorientation..."
Instructions:
1. Extract key reasoning steps.
2. Provide structured domain-specific assessment (Memory, Orientation, etc.).
3. Output JSON:
{
  "reasoning_steps": [...],
  "assessment": "...",
  "cdr_score": "[choose from 0.5, 1]"
}
\end{lstlisting}

We developed a four-stage CoT integrated diagnostic framework designed to simulate the collective reasoning process of expert clinical panels. This architecture overcomes the limitations of traditional single-threaded CoT prompts by integrating mechanisms for reasoning diversity, information fusion, logical calibration, and auditability. The entire system aims to enhance the interpretability and reliability of CDR grading, with the overall workflow illustrated in Figure \ref{fig:placeholder} and List \ref{lst:prompt_structure}.

The foundation of this framework is the CoT Generation and Diversity stage, serving as the core reasoning layer. For each patient record composed of subjective clinical notes (S) and a predefined binary CDR label pair (e.g., [0.5, 1.0]), the system initiates four independent reasoning processes. Each process is executed by a fine-tuned large language model. The model generates structured JSON-formatted reasoning outputs covering six diagnostic domains—memory, orientation, judgment and problem-solving, community affairs, family and hobbies, personal care—along with independent preliminary CDR scores. To ensure reasoning diversity and mitigate potential biases from deterministic reasoning, each reasoning attempt employs an independent random seed. This deliberate randomness, termed seed cracking, fosters diverse interpretive perspectives while maintaining consistency in the medical reasoning context. The system instantly rejects any JSON outputs with formatting errors or invalid structures, ensuring only complete analyses proceed to the next stage. This phase yields four fully independent, semantically rich diagnostic reports, each representing a unique reasoning pathway.

These four assessment results are treated equally. All evaluation texts are subsequently fed into a language model, which synthesizes the texts into a single coherent diagnostic narrative from a physician's perspective. This process integrates recurring clinical clues while resolving contradictions. The final CDR grade is determined by classifying this consolidated narrative. This means the final decision originates from the interpretation of the integrated text, rather than a direct aggregation of the initial scores.




The Final Classification and Logical Consistency Check stage transforms the integrated assessment into an authoritative diagnostic judgment. Here, the model assumes the role of a CDR scoring expert, generating final scores based on predefined label sets. To ensure robustness and data integrity, the system programmatically extracts predicted scores from model text responses using strict regular expression patterns. Should the model generate out-of-range values (e.g., 0.0 or 2.0), the system automatically executes a clamping operation, adjusting the score to the nearest valid label (e.g., [0.5, 1.0]). This mechanism mitigates rare yet potentially disruptive prediction errors while preserving the integrity of valid reasoning chains.

\begin{figure*}
  \centering
  \includegraphics[width=\linewidth]{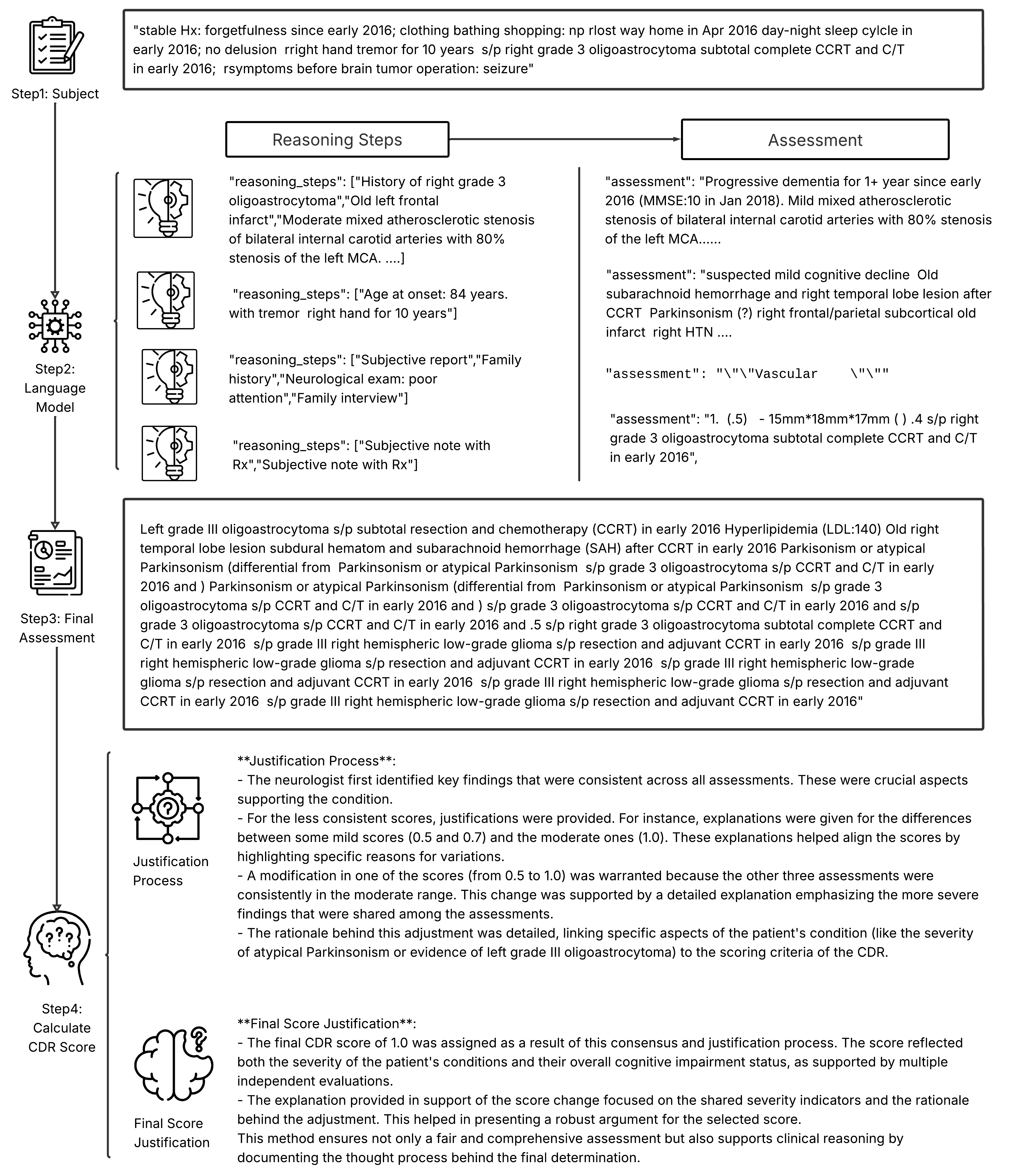}
  \caption{A CoT AI Workflow for Clinical Dementia Rating (CDR)  Assessment}
  \label{fig:pipeline}
\end{figure*}

In the final stage, summary Generation and Auditability Enhances interpretability by translating complex multi-stage reasoning into transparent, reviewable narratives. The system re-engages the model as a senior clinical consultant, synthesizing four initial CoT reports, evaluations, and final classification outcomes to summarize the entire diagnostic process. The generated textual audit trail provides clinicians with clear visibility into model reasoning transparency, ensuring each diagnostic conclusion is traceable to its underlying evidence, with the overall output example illustrated in Figure \ref{fig:pipeline}.

In summary, this four-phase CoT integration framework transforms traditional black-box LLM reasoning into a transparent, verifiable diagnostic process. By integrating reasoning diversity, structured aggregation, logical calibration, and explicit reasoning, the system achieves high robustness and interpretability, establishing itself as a reliable and explainable clinical AI solution for Alzheimer's disease assessment.

\section{Result}

Under the one-versus-one binary classification strategy, we conducted a comprehensive evaluation of model performance to reveal the actual capabilities of different models in distinguishing subtle CDR levels. 
Table \ref{tab:model_performance_2} compares zero-shot and fine-tuned models based on large language models for CoT diagnostic system.

\begin{table*}[htbp]
\centering
\caption{Performance of CoT-Based Large Language Models in Binary Classification Tasks}
\label{tab:model_performance_2}
\begin{tabular}{llccccc}
\toprule
\textbf{CDR Group} & \textbf{Models} & \textbf{Precision} & \textbf{Recall} & \textbf{F1-score} & \textbf{Accuracy} & \textbf{AUC} \\
\midrule
\multirow{4}{*}{0.5 vs. 1} 
 & QWEN2-7B zero shot prompting & 0.55 & 0.49 & 0.39 & 0.58 & 0.51 \\
 & Microsoft Phi-3B (CoT) & 0.23 & 0.50 & 0.32 & 0.46 & 0.50 \\
 & QWEN3-4B (CoT) & 0.43 & 0.45 & 0.42 & 0.49 & 0.45 \\
 & QWEN2-7B (CoT) & \textbf{0.60} & \textbf{0.56} & \textbf{0.54} & \textbf{0.61} & \textbf{0.56} \\

\midrule
\multirow{4}{*}{0.5 vs. 2}
 & QWEN2-7B zero shot prompting & 0.64 & 0.53 & 0.42 & 0.54 & 0.53 \\
 & Microsoft Phi-3B (CoT) & 0.59 & \textbf{0.58} & \textbf{0.56} & \textbf{0.59} & \textbf{0.58} \\
 & QWEN3-4B (CoT) & 0.58 & 0.53 & 0.45 & 0.56 & 0.53 \\
 & QWEN2-7B (CoT) & 0.29 & 0.45 & 0.35 & 0.54 & 0.45 \\

 \midrule
\multirow{4}{*}{0.5 vs. 3}  
 & QWEN2-7B zero shot prompting & 0.62 & 0.54 & 0.41 & 0.47 & 0.54 \\
 & Microsoft Phi-3B (CoT) & 1.00 & 0.33 & 0.50 & 0.33 & NaN \\
  & QWEN3-4B (CoT) & \textbf{0.72} & \textbf{0.55} & 0.40 & 0.47 & \textbf{0.55} \\
  & QWEN2-7B (CoT) & 0.54 & 0.54 & \textbf{0.53} & \textbf{0.54} & 0.54 \\
  
\midrule
\multirow{4}{*}{1 vs. 3}  
 & QWEN2-7B zero shot prompting   & 0.50 & 0.50 & 0.44 & 0.44 & 0.50 \\
 & Microsoft Phi-3B (CoT)   & 0.56 & 0.56 & \textbf{0.53} & \textbf{0.53} & 0.56 \\
 & QWEN3-4B (CoT)   &0.46  &0.47  &0.32  &0.32  &0.47  \\
 & QWEN2-7B (CoT)   & \textbf{0.58} & \textbf{0.57} & 0.50 & 0.50 & \textbf{0.57} \\
\bottomrule
\end{tabular}
\end{table*}

\begin{figure}
  \centering
  \includegraphics[width=\linewidth]{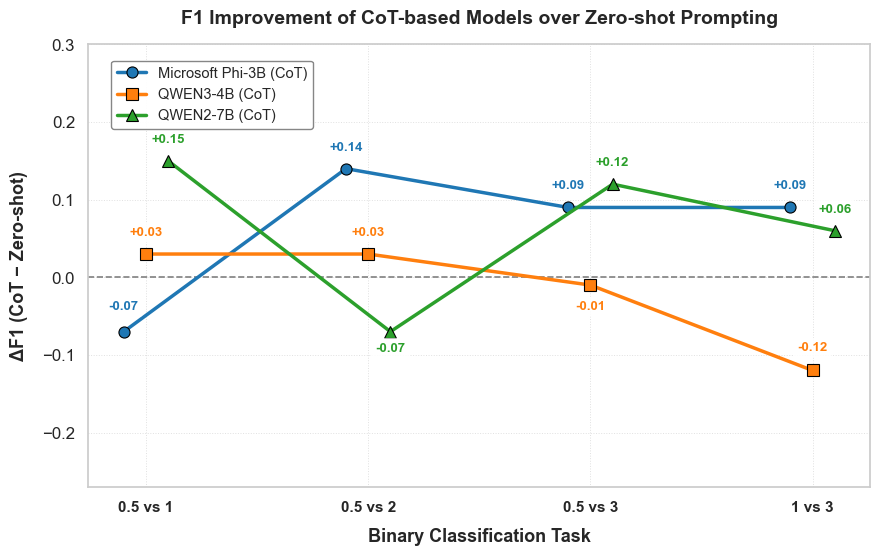}
  \caption{F1 Score Improvement of CoT-Based Models over Zero-Shot Prompting}
  \label{fig:f1}
\end{figure}

To evaluate the effectiveness of the proposed CoT framework, we conducted a comprehensive comparison across multiple LLMs under two distinct settings: standard zero-shot prompting, involving single-shot inference without explicit reasoning steps, and CoT-enhanced reasoning, involving stepwise intermediate reasoning. Experimental results are presented in Table \ref{tab:model_performance_2} and F1 scores across different models are shown in Figure \ref{fig:f1}. Findings demonstrate significant performance gains through CoT integration, particularly in distinguishing binary classification tasks across different CDR levels.

The CoT-based diagnostic system outperformed traditional prompting methods across most evaluation metrics, achieving superior F1 scores, accuracy, and AUC values. For instance, in the 0.5 vs. 1.0 classification task, the Qwen2-7B (CoT) model achieved an F1 score of 0.54 and accuracy of 0.61, outperforming the zero-shot version (F1 = 0.39, accuracy = 0.58) by +0.15 F1 score, demonstrating the effectiveness of explicit reasoning in fine-grained clinical differentiation. Similarly, the Qwen3-4B (CoT) model showed substantial improvement in fine-grained discrimination tasks like 0.5 vs. 2 (F1 = 0.45, accuracy = 0.56). These results indicate that structured reasoning effectively enhances models' ability to distinguish borderline dementia stages. The Qwen2-7B (CoT) model exhibits the most stable performance characteristics, maintaining balanced precision and recall across all task pairs. Notably, both metrics exceed 0.57 in the 1 vs 3 classification task. This stability indicates the model's ability to handle broader CDR variations while preserving decision reliability. In contrast, the Qwen3-4B (CoT) model performs stronger on moderate classification tasks (e.g., 0.5 vs 2) but shows declining performance on more distant class pairs (e.g., 1 vs 3), suggesting that its smaller parameter size may limit its generalization ability across varying degrees of severity.

In contrast, the Phi-3B (CoT) model exhibits significant performance fluctuations, revealing its limitations in complex diagnostic reasoning tasks. Although the model achieves competitive recall rates in certain subtasks, its precision and AUC values fluctuate dramatically, indicating unstable decision boundaries and inconsistent reasoning chains. For instance, the Phi-3B model achieved an F1 score of 0.56 in the 0.5 vs 2 task, yet its performance plummeted to 0.50 in the 0.5 vs 3 task. These inconsistencies suggest that small-scale models may struggle to maintain logical coherence across multi-step reasoning sequences, highlighting the need for sufficient model capacity to ensure the stability of CoT-based reasoning in medical diagnostic scenarios. 

Analysis of cross-model F1 score trends reveals model-scale effects and algorithmic influences. Qwen2-7B (CoT) consistently maintains the highest and most stable F1 score performance across all CDR tasks, indicating that greater model capacity enables construction of more coherent reasoning chains and enhances diagnostic discrimination capabilities. In contrast, Qwen3-4B (CoT) shows moderate gains at moderate discrepancy levels (0.5 vs 2) but exhibits weaker stability on distant category pairs (1 vs 3), indicating limited generalization in scenarios with severe cognitive discrepancies. Meanwhile, Phi-3B (CoT) exhibits fluctuating F1 scores, reflecting the sensitivity of smaller architectures to reasoning depth and task complexity. These trends collectively demonstrate that parameter scale and CoT prompt design synergistically enhance reasoning fidelity and stability. Notably, models exceeding 7 billion parameters appear to strike a critical balance between reasoning diversity and consistency, yielding more reliable diagnostic outputs.

Simultaneously, we observed the performance of multi-stage CoT frameworks on heterogeneous CDR classification tasks. Although absolute F1 scores fluctuated with task difficulty, CoT-enhanced models (particularly Qwen2-7B) avoided failure and maintained balanced precision-recall curves across all four binary settings. In contrast, the smaller Phi-3B (CoT) model exhibited significant instability, indicating poor discrimination capabilities under the most challenging scenarios. These results demonstrate that multi-stage reasoning structures, when combined with sufficient model capacity, can mitigate extreme performance fluctuations and maintain consistent decision boundaries across both adjacent and distant CDR comparisons.

CoT reasoning enhances diagnostic consistency and interpretability. Unlike traditional LLM prompts relying solely on implicit statistical correlations, CoT-based reasoning generates transparent intermediate reasoning paths that mirror clinicians' cognitive processes during dementia assessments. Furthermore, the multi-stage in CoT reasoning mitigates output instability by enforcing internal logical consistency between reasoning steps. In contrast, zero-shot prompts lacking CoT guidance often yield inconsistent and less reliable predictions due to the absence of structured reasoning supervision.


\section{Discussion}


\textcolor{black}{Clinical dementia assessments based on electronic health records rely on interpreting cognitive and functional behavioral descriptions recorded in routine clinical practice. These descriptions are often heterogeneous, implicit, and context-dependent, posing challenges for automated evaluation. In such scenarios, effective assessment depends not only on predictive accuracy but also on the stability of reasoning processes, the transparency of intermediate judgements, and the ability to trace diagnostic conclusions back to observable clinical evidence. To address clinical behavioral assessment, this study employs a structured, multi-stage CoT reasoning process to organize intermediate inferences and support consistent interpretation of clinical narratives. This design enables systematic analysis of complex behavioral information through explicit, auditable reasoning steps.}

Experimental results validate the effectiveness of integrating CoT reasoning into large language models for Alzheimer's disease assessment. Compared to existing diagnostic frameworks based on CoT or reasoning, this system focuses on structured electronic health record analysis rather than speech or synthetic datasets \cite{park2025reasoning}, providing evaluations with greater clinical evidence value. Furthermore, unlike general medical reasoning studies, this model is specifically optimized for Alzheimer's disease staging and validated across diverse patient datasets, demonstrating robustness and scalability. These findings highlight the practical application potential of LLM-based diagnostic systems for real-world dementia assessment while maintaining transparency and clinical interpretability. Furthermore, unlike studies on general medical reasoning, our model is specifically optimized for Alzheimer's disease staging and validated across diverse patient datasets, demonstrating robustness and scalability \cite{li2025care}. From a clinical perspective, the enhancement of reasoning stability increases clinician confidence, reduces diagnostic variability among raters, and thereby strengthens the interpretability and reliability of AI-assisted assessments. These findings highlight the practical potential of LLM-based diagnostic systems for real-world dementia evaluation while maintaining transparency and clinical interpretability.


This study demonstrates that integrating CoT reasoning into large language models significantly enhances their diagnostic stability and interpretability in CDR assessments. Beyond numerical performance gains, the CoT framework transforms conventional large language models into structured reasoning systems capable of articulating clinical evidence, weighing diagnostic clues, and integrating multi-perspective judgments. Consistent improvements observed across both Qwen2-7B and Qwen3-4B models indicate that CoT integration offers benefits independent of specific model architectures.

The CoT-based diagnostic system introduces a transparent reasoning mechanism aligned with clinical expert decision-making processes. By explicitly generating intermediate reasoning paths and employing multi-stage evaluation, this system transforms the previously opaque prediction process into a traceable, verifiable chain of reasoning, significantly enhancing the model's credibility in medical applications. Simultaneously, the multi-agent design effectively mitigates the randomness inherent in large language models during reasoning, leading to more stable performance on complex borderline tasks.

It should be noted that while CoT reasoning incurs additional computational overhead (due to multiple inference steps), this trade-off is acceptable as the resulting gains in diagnostic reliability and interpretability hold significant value for clinical applications. This approach bridges automated prediction with explainable AI, providing a scalable technical foundation for future deployment in real-world medical settings.


Although these findings are encouraging, the study has limitations. The dataset size is relatively small, and the analysis relies solely on text-based electronic health records (EHRs), excluding imaging data such as magnetic resonance imaging (MRI) or positron emission tomography (PET). Furthermore, external validation using independent cohorts is still required to confirm its generalizability. Future research will explore multimodal integration (e.g., combining EHR and MRI features) and human-machine collaborative validation frameworks to enhance interpretability and clinical reliability.
\section{Conclusion}
This study proposes a CoT reasoning framework based on large language models  for the clinical assessment and diagnosis of AD. Unlike traditional black-box fine-tuning approaches, this system explicitly simulates the clinical physician's reasoning process by generating intermediate diagnostic inferences and multi-layer evaluations. Experimental results demonstrate that the proposed CoT diagnostic system not only significantly enhances model interpretability and decision transparency but also exhibits higher consistency and stability in distinguishing between adjacent CDR grades. Enhancing reasoning transparency not only improves interpretability at the individual diagnostic level but also establishes a scalable framework for trustworthy clinical artificial intelligence. This advancement facilitates the integration of interpretable large language models into healthcare systems.

By transforming the model's reasoning process into a structured, traceable chain of reasoning, this study establishes a novel connection between automated prediction and explainable artificial intelligence, providing a scalable technical foundation for large language models in clinical decision support systems.

Future research will further expand the CoT reasoning framework and integrate longitudinal EHR information to achieve multimodal diagnostic fusion. This will enable a more comprehensive characterization of disease dynamics and advance precision medicine applications in dementia diagnosis. Additionally, external validation using multicenter datasets will be conducted, alongside exploring the design of human-machine collaborative reasoning systems, laying the groundwork for practical deployment in clinical settings.

\bibliographystyle{IEEEtran}
\bibliography{ref}

\end{document}